\pdfoutput=1

\documentclass[11pt]{article}

\usepackage[]{acl}

\usepackage{times}
\usepackage{latexsym}
\usepackage{multirow}

\usepackage[T1]{fontenc}

\usepackage[utf8]{inputenc}

\usepackage{microtype}

\usepackage{graphicx}
\usepackage{arydshln}

\title{Building Multilingual Machine Translation Systems \\That Serve Arbitrary \emph{X-Y} Translations}

\author{
    Akiko Eriguchi$^{*\,\dagger}$, 
    Shufang Xie\thanks{\ \ Equal contributions.} $^{\ \ddagger}$\ ,
    Tao Qin$^{\ddagger}$, \and Hany Hassan Awadalla$^{\dagger}$ \\
    $^{\dagger}$Microsoft \ \ \ $^{\ddagger}$Microsoft Research Asia \\
    \texttt{\{akikoe,shufxi,taoqin,hanyh\}@microsoft.com}
}

\newcommand{\basec}{12E6D}
\newcommand{\bigc}{24E12D}

\begin{document}
\maketitle
\begin{abstract}

Multilingual Neural Machine Translation (MNMT) enables one system to translate sentences from multiple source languages to multiple target languages, greatly reducing deployment costs compared with conventional bilingual systems.
The MNMT training benefit, however, is often limited to many-to-one directions. The model suffers from poor performance in one-to-many and many-to-many with zero-shot setup.
To address this issue, this paper discusses how to practically build MNMT systems that serve arbitrary \texttt{X-Y} translation directions while leveraging multilinguality with a two-stage training strategy of pretraining and finetuning. 
Experimenting with the WMT'21 multilingual translation task, we demonstrate that our systems outperform the conventional baselines of direct bilingual models and pivot translation models for most directions, averagely giving +6.0 and +4.1 BLEU, without the need for architecture change or extra data collection.
Moreover, we also examine our proposed approach in an extremely large-scale data setting to accommodate practical deployment scenarios.  
\end{abstract}

\section{Introduction}
Multilingual Neural Machine Translation (MNMT), which enables one system to serve translation for multiple directions, has attracted much attention in the machine translation area~\cite{zoph-knight-2016-multi,firat-etal-2016-zero}. Because the multilingual capability hugely reduces the deployment cost at training and inference, MNMT has actively been employed as a machine translation system backbone in recent years~\cite{johnson-etal-2017-googles,DBLP:journals/corr/abs-1803-05567}.

Most MNMT systems are trained with multiple English-centric data for both directions (e.g., English $\rightarrow$ \{French, Chinese\} (\texttt{En-X}) and \{French, Chinese\} $\rightarrow$ English (\texttt{X-En})). Recent work~\cite{gu-etal-2019-improved,zhang-etal-2020-improving,yang-etal-2021-improving-multilingual} pointed out that such MNMT systems severely face an off-target translation issue, especially in translations from a non-English language X to another non-English language Y. Meanwhile, \citet{freitag-firat-2020-complete} have extended data resources with multi-way aligned data and reported that one complete many-to-many MNMT can be fully supervised, achieving competitive translation performance for all \texttt{X-Y} directions. In our preliminary experiments, we observed that the complete many-to-many training is still as challenging as one-to-many training~\cite{johnson-etal-2017-googles,wang-etal-2020-negative}, since we have introduced more one-to-many translation tasks into the training. Similarly reported in the many-to-many training with zero-shot setup \cite{gu-etal-2019-improved,yang-etal-2021-improving-multilingual}, the complete MNMT model also suffers from capturing correlations in the data for all the \texttt{X-Y} directions as one model training, due to highly imbalanced data. 

\begin{figure*}[t]
    \includegraphics[width=\linewidth]{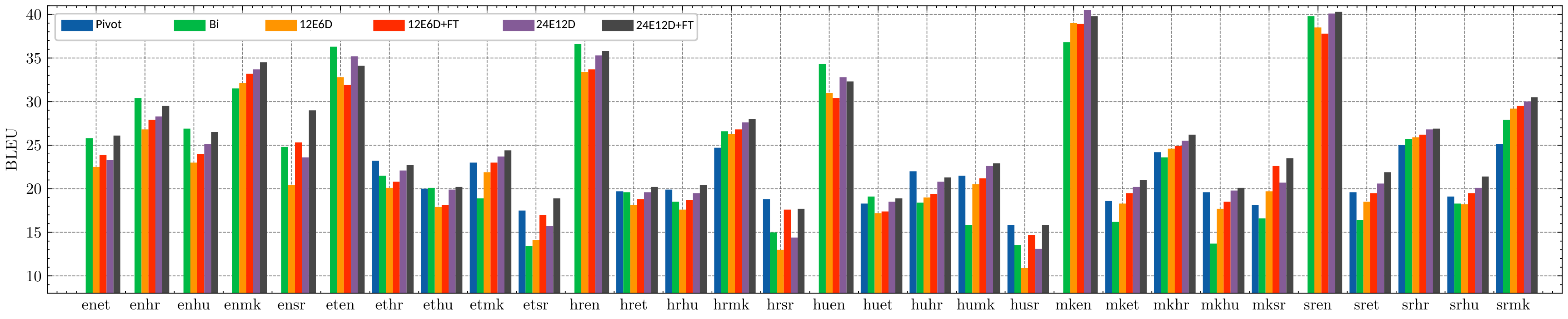}
    \includegraphics[width=\linewidth]{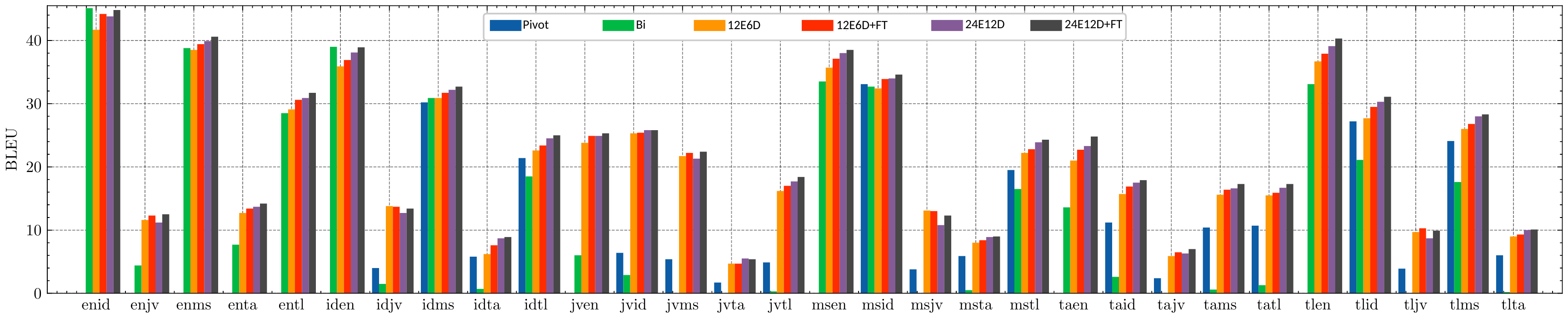}
    \caption{Average translation performance of our systems in the WMT'21 large-scale multilingual translation Task 1 (\textbf{top}) and Task 2 (\textbf{bottom}), with the respective average improvement of (\basec, \basec+FT, \bigc, \bigc+FT) = (+3.6, +4.7, +5.0, \textbf{+6.0}) and (+2.0, +2.9, +3.2, \textbf{+4.1}) against the bilingual baseline (\lq\lq Bi'') and the pivot translation baselines (\lq\lq Pivot''). \lq\lq \basec /\bigc'' denote our two settings, with \lq\lq+FT'' suffix for finetuned systems.
    }
    \label{fig:results}
\end{figure*}
In this paper, we propose a two-stage training for complete MNMT systems that serve arbitrary \texttt{X-Y} translations by 1) pretraining a complete multilingual many-to-many model and 2) finetuning the model to effectively transfer knowledge from pretraining to task-specific multilingual systems. Considering that MNMT is a multi-task learner of translation tasks with \lq\lq multiple languages'', the complete multilingual model learns more diverse and general multilingual representations. We transfer the representations to a specifically targeted task via many-to-one multilingual finetuning, and eventually build multiple many-to-one MNMT models that cover all \texttt{X-Y} directions. The experimental results on the WMT'21 multilingual translation task show that our systems have substantial improvement against conventional bilingual approaches and many-to-one multilingual approaches for most directions. Besides, we discuss our proposal in the light of feasible deployment scenarios and show that the proposed approach also works well in an extremely large-scale data setting.

\section{Two-Stage Training for MNMT Models}~\label{sec: multilingual nmt}

To support all possible translations with $|L|$ languages (including English), we first train a complete MNMT system on all available parallel data for $|L| \times (|L|-1)$ directions. 
We assume that there exist data of $(|L|-1)$ English-centric language pairs and remaining $\frac{ (|L|-1)\times (|L|-2)}{2}$ non-English-centric language pairs, which lets the system learn multilingual representations across all $|L|$ languages.
Usually, the volume of English-centric data is much greater than non-English-centric one.
Then, we transfer the multilingual representations to one target language $L$ by finetuning the system on a subset of training data for many-to-$L$ directions (i.e., \textbf{multilingual many-to-one finetuning}). This step leads the decoder towards the specifically targeted language $L$ rather than multiple languages. As a result, we obtain $|L|$ multilingual many-to-one systems to serve all \texttt{X-Y} translation directions. We experiment with our proposed approach in the following two settings: 1) WMT'21 large-scale multilingual translation data with 972M sentence pairs and 2) our in-house production-scale dataset with 4.1B sentence pairs.

\begin{table*}[th]
\centering
\begin{tabular}{lccccccccccccc}
\hline
\textbf{model} & \textbf{en} & \textbf{hu}& \textbf{hr} & \textbf{sr} & \textbf{et} & \textbf{id}  & \textbf{ms} & \textbf{tl} & \textbf{mk} & \textbf{jv} & \textbf{ta}  & \textbf{avg.} \\
\hline

Bilingual& {30.4} & {19.5}  & {23.9} & {16.7} & {19.4} & {20.9} & {17.6} & {13.0} & {24.1} & {1.2} & {1.8} & {17.1}\\
Pivot-based & {30.4} & {19.7} & {23.6} & {17.6} & {19.1} & {19.5} & {21.8} & {14.1} & {23.6} & {3.5} & {5.4} & {18.0}\\
Many-to-one & {32.9} & \textbf{20.8} & \textbf{24.0} & {17.4} & \textbf{19.8} & {25.3} & {20.3} & {16.6} & {25.0} & {1.5} & {5.1} & {19.0}\\ 
\hdashline
Ours: Pretrained & {32.0} & {18.9} & {23.4} & {15.6} & {19.0} & {28.6} & {26.5} & {21.1} & {26.2} & {10.8}  & {8.1} & {20.9}\\
 + finetuning & \textbf{33.2} & {19.8} & {23.8} & \textbf{19.5} & \textbf{19.8} & \textbf{30.0} & \textbf{27.3} & \textbf{21.9} & \textbf{26.8} & \textbf{11.2} & \textbf{8.7} & \textbf{22.0}\\
 \hline \hline
 Data size (M) & 321 & 172 & 141 & 123 & 85 & 63 & 20 & 19 & 18 & 5 & 4 & -- \\
\hline
\end{tabular}
\caption{Average sacreBLEU scores for many-to-$L$ directions on both Task 1 and 2, and the data statistics of the corresponding $L$-centric training data  ($L$=\{en, hu, hr, sr, et, id, ms, tl, mk, jv, ta\}). All the multilingual systems including many-to-one baselines and the proposed model are \basec. Note that the \lq\lq Pivot-based" system for many-to-English directions is identical to \lq\lq Bilingual". }
\label{tab:result_wmt}
\end{table*}

\section{WMT'21 Multilingual Translation Task} \label{sec: wmt21}
We experiment with two small tasks of the WMT'21 large-scale multilingual translation task.
The tasks provide multilingual multi-way parallel corpora from the Flores 101 data~\cite{wenzek2021findings}.
The parallel sentences are provided among English (en), five Central and East European languages of \{Croatian (hr), Hungarian (hu), Estonian (et), Serbian (sr), Macedonian (mk)\} for the task 1, and five Southeast Asian languages of \{Javanese (jv), Indonesian (id), Malay (ms), Tagalog (tl), Tamil (ta)\} for the task 2. We removed sentence pairs either of whose sides is an empty line, and eventually collected the data
with (English-centric, Non-English-centric)=(321M, 651M) sentence pairs in total. The data size per direction varies in a range of 0.07M-83.9M. To balance the data distribution across languages~\cite{kudugunta-etal-2019-investigating}, we up-sample the low-resource languages with temperature=5.
We append language ID tokens at the end of source sentences to specify a target language~\cite{johnson-etal-2017-googles}. We tokenize the data with the SentencePiece~\cite{kudo-richardson-2018-sentencepiece} and build a shared vocabulary with 64k tokens.

We train Transformer models~\cite{NIPS2017_3f5ee243} consisting of a $m$-layer encoder and $n$-layer decoder with (hidden dim., ffn dim.) =(768, 3072) in a complete multilingual many-to-many fashion. We have two settings of ($m$, $n$) = (12, 6) for \lq\lq \basec'' and (24, 12) for \lq\lq\bigc'' , to learn diverse multilingual dataset. 
The model parameters are optimized by using RAdam~\cite{Liu2020On} with an initial learning rate of 0.025, and warm-up steps of 10k and 30k for the \basec~and \bigc~model training, respectively. The systems are pretrained on 64 V100 GPUs with a mini-batch size of 3072 tokens and gradient accumulation of 16. After the pretraining, the models are finetuned on a subset of \texttt{X-}$L$ training data.
We finetune the model parameters gently on 8 V100 GPUs with the same mini-batch size, gradient accumulations, and optimizer with different learning rate scheduling of (init\_lr, warm-up steps)=(\{1e-4, 1e-5, 1e-6\}, 8k). The best checkpoints are selected based on development loss. The translations are obtained by a beam search decoding with a beam size of 4, unless otherwise stated.

\paragraph{Baselines}
For system comparison, we build three different baselines: 1) direct bilingual systems, 2) pivot translation systems via English (only applicable for non-English \texttt{X-Y} evaluation) \cite{utiyama-isahara-2007-comparison}, and 3) many-to-one multilingual systems with the \basec~ architecture. The bilingual and pivot-based baselines employ the Transformer base architecture. 
The embedding dimension is set to 256 for jv, ms, ta, and tl, because of the training data scarcity.
For the \texttt{X-Y} pivot translation, a source sentence in language \texttt{X} is translated to English with a beam size of 5 by the \texttt{X-En} model, then the best output is translated to the final target language \texttt{Y} by the \texttt{En-Y} model.

\paragraph{Results} 
All results on the test sets are displayed in Figure~\ref{fig:results} and Table~\ref{tab:result_wmt}, where we report the case-sensitive sacreBLEU score~\cite{post-2018-call} for translation accuracy. Overall, our best systems (\lq\lq \bigc+FT") are significantly better by $\geq +0.5$ sacreBLEU for 83\% and 88\% directions against the bilingual baselines and the pivot translation baselines, respectively. In Table~\ref{tab:result_wmt}, we present the average sacreBLEU scores for many-to-$L$ directions, showing that our proposed approach successfully achieved the best performance in most targeted languages. Compared to the many-to-one multilingual baselines, the proposed approach of utilizing the complete MNMT model transfers multilingual representations more effectively to the targeted translation directions, as the $L$-centric data size are smaller. We also note that the winning system of the shared task achieved (task1, task2)=(37.6, 33.9) BLEU with a 36-layer encoder and 12-layer decoder model~\cite{yang-etal-2021-multilingual-machine} that is pre-trained on extra language data including parallel and monolingual data, while our best system with a 24-layer encoder and 12-layer decoder obtained (task1, task2)=(25.7, 22.8) sacreBLEU, without using those extra data.

\begin{table*}[!th]
\centering
\begin{tabular}{lccccc}
\hline
\textbf{} & \textbf{xx-de} & \textbf{xx-fr} & \textbf{xx-es} & \textbf{xx-it} & \textbf{xx-pl} \\
\hline
Pivot-based baselines    & {35.7}   & {39.8}   & {38.0}   & {33.0}   & {26.2} \\
Pretrained model on en-xx data    & {14.2}   & {33.1}   & {---}   & {---}   & {---} \\
 + Many-to-one multilingual finetuning & {36.6}   & {41.3}   & {---}   & {---}   & {---} \\
Pretrained model on \{en, de, fr\}-xx data & {37.5}   & {42.1}   & {20.0}   & {16.6}   & {11.1} \\
 + Many-to-one multilingual finetuning & \textbf{38.3} & \textbf{42.6} & \textbf{39.3} & \textbf{34.6} & \textbf{28.3} \\\hline
\end{tabular}
\caption{Average sacreBLEU scores of the proposed two-stage MNMT training for the many-to-one directions, with the English-centric pretrained model and the multi-centric model pretrained on \{en, de, fr\}-xx data. }
\label{tab:result_finetune}
\end{table*}
\section{In-house Extremely Large-Scale Setting}
Deploying a larger and larger model is not always feasible. We often have limitations in the computational resources at inference time, which leads to a trade-off problem between the performance and the decoding cost caused by the model architecture. In this section, we validate our proposed approach in an extremely large-scale data setting and also discuss how we can build lighter NMT models without the performance loss, while distilling the proposed MNMT systems \cite{kim-rush-2016-sequence}.
We briefly touch the following three topics of 1) multi-way multilingual data collection, 2) English-centric vs. multi-centric pretraining for \texttt{X-Y} translations, and 3) a lighter NMT model that addresses the trade-off issue between performance and latency. Then, we report the experimental results in the extremely large-scale setting.

\paragraph{Multilingual Data Collection}
We build an extremely large-scale data set using our in-house English-centric data set, consisting of 10 European languages, ranging 24M-192M sentences per language. This contains available parallel data and back-translated data between English and \{German (de), French (fr), Spanish (es), Italian (it), Polish (pl), Greek (el), Dutch (nl), Portuguese (pt), and Romanian (ro)\}. From these English-centric corpora, we extract a  multi-way multilingual \texttt{X-Y} data, by aligning \texttt{En-X} and \texttt{En-Y} data via pivoting English. Specifically, we extracted \{de, fr, es, it, pl\}-centric data and concatenate them to the existent direct \texttt{X-Y} data, providing 78M-279M sentence pairs per direction. 
Similarly as in Section~\ref{sec: wmt21}, we build a shared SentencePiece vocabulary with 128k tokens to address the large-scale setting.

\paragraph{En-centric vs Multi-centric Pretraining} In a large-scale data setting, a question might come up; \emph{Which pretrained model provides generalized multilingual representations to achieve better \texttt{X-Y} translation quality?} Considering English is often a dominant text data, e.g., 70\% tasks are English-centric in the WMT'21 news translation task, the model supervised on English-centric corpora might learn representations enough to transfer for \texttt{X-Y} translations. To investigate the usefulness of the multi-centric data training, we pretrain our Transformer models with deeper 24-12 layers described in Section~\ref{sec: wmt21}, on the English-centric data and the $L$-centric data ($L$=\{en,de,fr\}), individually. After pretraining, we apply the multilingual many-to-one finetuning with a subset of the training data and evaluate each system for the fully supervised \texttt{X-Y} directions, i.e., \texttt{xx}-\{\texttt{en,de,fr}\}, and the partially supervised \texttt{X-Y} directions, i.e., \texttt{xx}-\{\texttt{es,it,pl}\}. We followed the same training and finetuning settings as described in Section~\ref{sec: wmt21}, unless otherwise stated.

\paragraph{MNMT with Light Decoder} \label{sec: light_model}
At the practical level, one drawback of the large-scale models would be latency at inference time. This is mostly caused by the high computational cost in the decoder layers due to auto-regressive models and the extra cross-attention network in each block of the decoder. Recent studies \cite{kasai2021deep,hsu-etal-2020-efficient,Li_Lin_Xiao_Zhu_2021} have experimentally shown that models with a deep encoder and a shallow decoder can address the issue, without losing much performance. Fortunately, such an architecture also satisfies demands of the many-to-one MNMT training, which requires the encoder networks to be more complex to handle various source languages. To examine the light NMT model architecture, we train the Transformer base architecture modified with 9-3 layers (\texttt{E9D3}) in a bilingual setting and compare it with a standard Transformer base model, with 6-6 layers  (\texttt{E6D6}), as a baseline. Additionally, we also report direct \texttt{X-Y} translation performance, when distilling the best large-scale MNMT models alongside the light NMT models as a student model. More specifically, following \citet{kim-rush-2016-sequence}, we train light NMT student models (\texttt{E9D3}) that serve many-to-$L$ translations ($L$=\{de, fr, es, it, pl\}).

\paragraph{Results}
Table~\ref{tab:result_finetune} reports average sacreBLEU scores for many-to-one directions in our in-house X-Y test sets. For the \texttt{xx-\{de,fr\}} directions, the proposed finetuning helps both English-centric and multi-centric pretrained models to improve the accuracy. Overall, the finetuned multi-centric models achieved the best, largely outperforming the English pivot-based baselines by +2.6 and +2.8 points. For the comparison among the multilingual systems, the multi-centric model without finetuning already surpasses the finetuned English-centric systems with a large margin of +0.9 and +0.8 points for both \texttt{xx-\{de,fr\}} directions. This suggests that, by pretraining a model on more multi-centric data, the model learns better multilinguality to transfer. For the \texttt{xx-\{es,it,pl\}} directions\footnote{Most are zero-shot directions such as \lq\lq Greek-to-Spanish''.}, the fineutned multi-centric systems gain similar accuracy improvement, averagely outperforming the conventional pivot-based baselines.

\begin{figure}[!tbp]
    \centering
    \includegraphics[width=\linewidth]{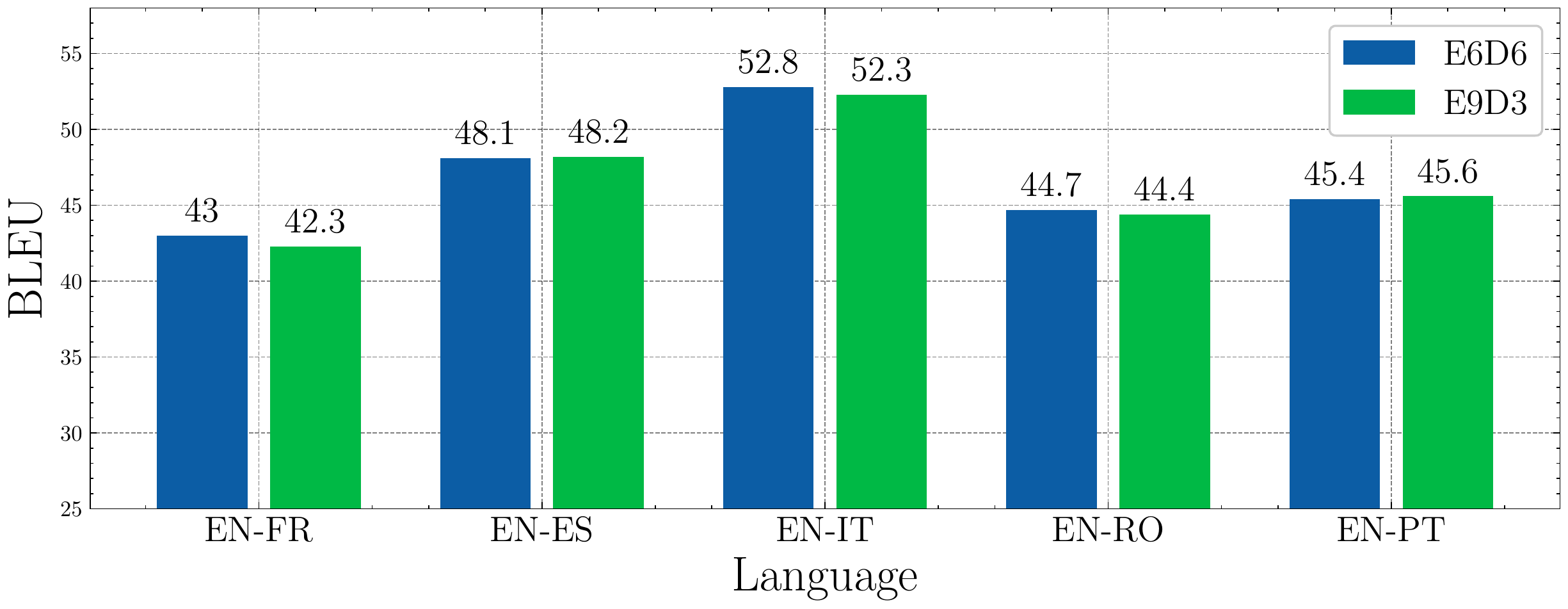}
    \includegraphics[width=\linewidth]{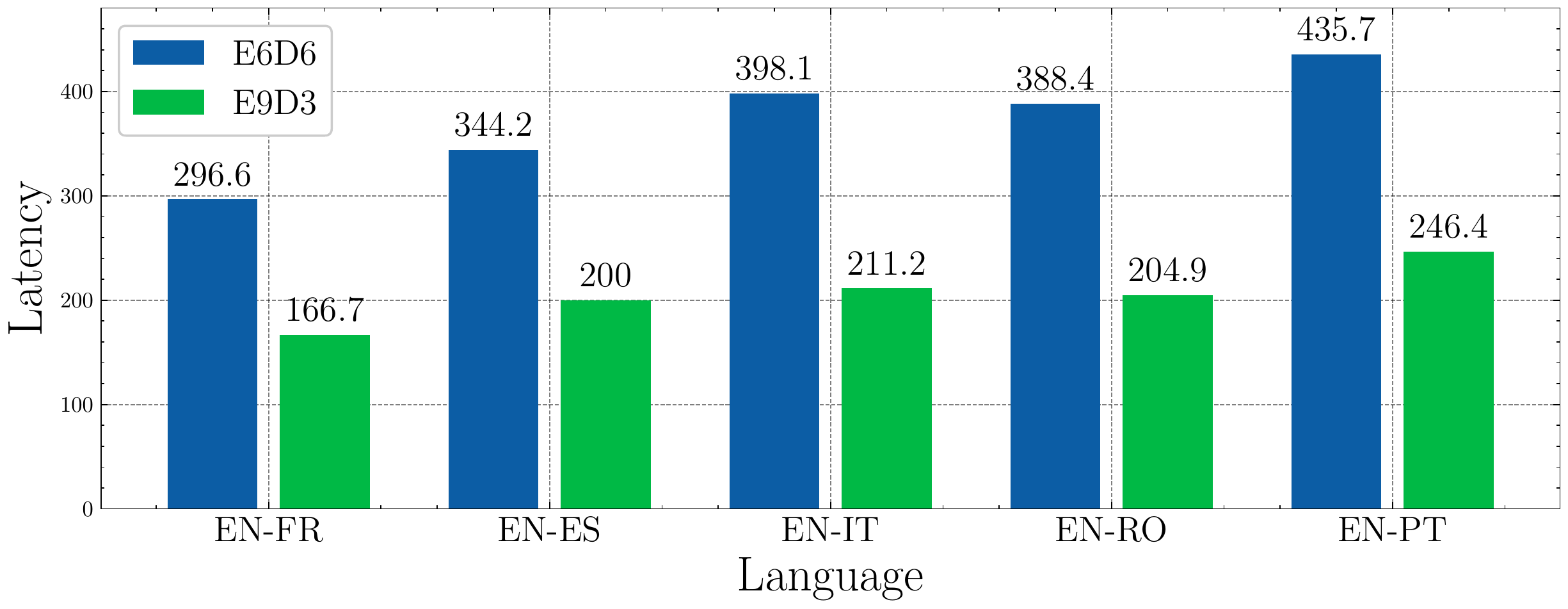}
    \caption{The BLEU score and latency in milliseconds of standard~(\texttt{E6D6}) and light~(\texttt{E9D3}) architecture.}
    \label{fig:light}
\end{figure}

Figure \ref{fig:light} shows the effectiveness of our light NMT model architecture for five bilingual \texttt{En-X} directions, reporting the translation performance in sacreBLEU scores and the latency measured on CPUs. Our light NMT model (\texttt{E9D3}) successfully achieves almost 2x speed up, without much drop of the performance for all directions. Employing this light model architecture as a student model, we report the distilled many-to-one model performance in Table~\ref{tab:result_stud}, measured by sacreBLEU and COMET scores~\cite{rei-etal-2020-comet}. For consistent comparison, we also built English bilingual baselines (\texttt{E6D6}) that are distilled from the bilingual Teachers, then we obtained the English pivot-based translation performance. For all the many-to-$L$ directions ($L$=\{de,fr,es,it,pl\}), the light NMT models that are distilled from the best MNMT models show the best performance in both metrics. 
Besides that, we also note that our direct \texttt{X-Y} light NMT systems successfully save the decoding cost with 75\% against the pivot translation\footnote{The light NMT model halves the latency against the baseline system, as shown in in Figure~\ref{fig:light} and needs to be run once. On the other hand, the pivot-based baseline systems via English need to translate twice for \texttt{X-Y} directions (e.g., German-to-English and English-to-French translations for a German-French direction).}.

\begin{table} [t]
\centering
\scalebox{0.97}{
\begin{tabular}{@{}clcc@{}}
\hline
& \textbf{Models} & \textbf{BLEU} & \textbf{COMET} \\
\hline
\multirow{2}{*}{\textbf{xx-de}} & Pivot-based baselines & {38.1} & {60.8} \\
& Ours & \textbf{39.5} & \textbf{67.2} \\
\hline
\multirow{2}{*}{\textbf{xx-fr}} & Pivot-based baselines & {41.5} & {69.3} \\
& Ours & \textbf{42.9} & \textbf{73.2} \\
\hline
\multirow{2}{*}{\textbf{xx-es}} &  Pivot-based baselines & {37.4} & {73.9} \\
& Ours & \textbf{38.0} & \textbf{74.4} \\
\hline
\multirow{2}{*}{\textbf{xx-it}} &  Pivot-based baselines & {32.6} & {77.2} \\
& Ours & \textbf{33.7} & \textbf{80.9} \\
\hline
\multirow{2}{*}{\textbf{xx-pl}} &  Pivot-based baselines & {26.7} & {77.8} \\
& Ours & \textbf{28.0} & \textbf{88.5}  \\\hline
\end{tabular}
}
\caption{Average direct \texttt{X-Y} translation performance of our proposed light NMT models. All \lq\lq Our'' NMT systems employ the light models (\texttt{E9D3}) that are distilled from the best systems reported in Table~\ref{tab:result_finetune}.}
\label{tab:result_stud}
\end{table}

\section{Conclusion}
This paper proposes a simple but effective two-stage training strategy for MNMT systems that serve arbitrary \texttt{X-Y} translations. To support translations across languages, we first pretrain a complete multilingual many-to-many model, then transfer the representations via finetuning the model in a many-to-one multilingual fashion. In the WMT'21 translation task, we experimentally showed that the proposed approach substantially improve translation accuracy for most \texttt{X-Y} directions against the strong conventional baselines of bilingual systems, pivot translation systems, and many-to-one multilingual systems. We also examined the proposed approach in the extremely large-scale setting, while addressing the practical questions such as multi-way parallel data collection, the usefulness of multilinguality during the pretraining and finetuning, and how to save the decoding cost, achieving the better \texttt{X-Y} quality.  


\bibliography{anthology,custom}

\begin{thebibliography}{22}
\expandafter\ifx\csname natexlab\endcsname\relax\def\natexlab#1{#1}\fi

\bibitem[{Firat et~al.(2016)Firat, Sankaran, Al-onaizan, Yarman~Vural, and
  Cho}]{firat-etal-2016-zero}
Orhan Firat, Baskaran Sankaran, Yaser Al-onaizan, Fatos~T. Yarman~Vural, and
  Kyunghyun Cho. 2016.
\newblock \href {https://doi.org/10.18653/v1/D16-1026} {Zero-resource
  translation with multi-lingual neural machine translation}.
\newblock In \emph{Proceedings of the 2016 Conference on Empirical Methods in
  Natural Language Processing}, pages 268--277, Austin, Texas. Association for
  Computational Linguistics.

\bibitem[{Freitag and Firat(2020)}]{freitag-firat-2020-complete}
Markus Freitag and Orhan Firat. 2020.
\newblock \href {https://aclanthology.org/2020.wmt-1.66} {Complete multilingual
  neural machine translation}.
\newblock In \emph{Proceedings of the Fifth Conference on Machine Translation},
  pages 550--560, Online. Association for Computational Linguistics.

\bibitem[{Gu et~al.(2019)Gu, Wang, Cho, and Li}]{gu-etal-2019-improved}
Jiatao Gu, Yong Wang, Kyunghyun Cho, and Victor~O.K. Li. 2019.
\newblock \href {https://doi.org/10.18653/v1/P19-1121} {Improved zero-shot
  neural machine translation via ignoring spurious correlations}.
\newblock In \emph{Proceedings of the 57th Annual Meeting of the Association
  for Computational Linguistics}, pages 1258--1268, Florence, Italy.
  Association for Computational Linguistics.

\bibitem[{Hassan et~al.(2018)Hassan, Aue, Chen, Chowdhary, Clark, Federmann,
  Huang, Junczys{-}Dowmunt, Lewis, Li, Liu, Liu, Luo, Menezes, Qin, Seide, Tan,
  Tian, Wu, Wu, Xia, Zhang, Zhang, and
  Zhou}]{DBLP:journals/corr/abs-1803-05567}
Hany Hassan, Anthony Aue, Chang Chen, Vishal Chowdhary, Jonathan Clark,
  Christian Federmann, Xuedong Huang, Marcin Junczys{-}Dowmunt, William Lewis,
  Mu~Li, Shujie Liu, Tie{-}Yan Liu, Renqian Luo, Arul Menezes, Tao Qin, Frank
  Seide, Xu~Tan, Fei Tian, Lijun Wu, Shuangzhi Wu, Yingce Xia, Dongdong Zhang,
  Zhirui Zhang, and Ming Zhou. 2018.
\newblock \href {http://arxiv.org/abs/1803.05567} {Achieving human parity on
  automatic chinese to english news translation}.
\newblock \emph{arXiv: arXiv:1803.05567}.

\bibitem[{Hsu et~al.(2020)Hsu, Garg, Liao, and
  Chatsviorkin}]{hsu-etal-2020-efficient}
Yi-Te Hsu, Sarthak Garg, Yi-Hsiu Liao, and Ilya Chatsviorkin. 2020.
\newblock \href {https://doi.org/10.18653/v1/2020.sustainlp-1.7} {Efficient
  inference for neural machine translation}.
\newblock In \emph{Proceedings of SustaiNLP: Workshop on Simple and Efficient
  Natural Language Processing}, pages 48--53, Online. Association for
  Computational Linguistics.

\bibitem[{Johnson et~al.(2017)Johnson, Schuster, Le, Krikun, Wu, Chen, Thorat,
  Vi{\'e}gas, Wattenberg, Corrado, Hughes, and
  Dean}]{johnson-etal-2017-googles}
Melvin Johnson, Mike Schuster, Quoc~V. Le, Maxim Krikun, Yonghui Wu, Zhifeng
  Chen, Nikhil Thorat, Fernanda Vi{\'e}gas, Martin Wattenberg, Greg Corrado,
  Macduff Hughes, and Jeffrey Dean. 2017.
\newblock \href {https://doi.org/10.1162/tacl_a_00065} {{G}oogle{'}s
  multilingual neural machine translation system: Enabling zero-shot
  translation}.
\newblock \emph{Transactions of the Association for Computational Linguistics},
  5:339--351.

\bibitem[{Kasai et~al.(2021)Kasai, Pappas, Peng, Cross, and
  Smith}]{kasai2021deep}
Jungo Kasai, Nikolaos Pappas, Hao Peng, James Cross, and Noah Smith. 2021.
\newblock \href {https://openreview.net/forum?id=KpfasTaLUpq} {Deep encoder,
  shallow decoder: Reevaluating non-autoregressive machine translation}.
\newblock In \emph{International Conference on Learning Representations}.

\bibitem[{Kim and Rush(2016)}]{kim-rush-2016-sequence}
Yoon Kim and Alexander~M. Rush. 2016.
\newblock \href {https://doi.org/10.18653/v1/D16-1139} {Sequence-level
  knowledge distillation}.
\newblock In \emph{Proceedings of the 2016 Conference on Empirical Methods in
  Natural Language Processing}, pages 1317--1327, Austin, Texas. Association
  for Computational Linguistics.

\bibitem[{Kudo and Richardson(2018)}]{kudo-richardson-2018-sentencepiece}
Taku Kudo and John Richardson. 2018.
\newblock \href {https://doi.org/10.18653/v1/D18-2012} {{S}entence{P}iece: A
  simple and language independent subword tokenizer and detokenizer for neural
  text processing}.
\newblock In \emph{Proceedings of the 2018 Conference on Empirical Methods in
  Natural Language Processing: System Demonstrations}, pages 66--71, Brussels,
  Belgium. Association for Computational Linguistics.

\bibitem[{Kudugunta et~al.(2019)Kudugunta, Bapna, Caswell, and
  Firat}]{kudugunta-etal-2019-investigating}
Sneha Kudugunta, Ankur Bapna, Isaac Caswell, and Orhan Firat. 2019.
\newblock \href {https://doi.org/10.18653/v1/D19-1167} {Investigating
  multilingual {NMT} representations at scale}.
\newblock In \emph{Proceedings of the 2019 Conference on Empirical Methods in
  Natural Language Processing and the 9th International Joint Conference on
  Natural Language Processing (EMNLP-IJCNLP)}, pages 1565--1575, Hong Kong,
  China. Association for Computational Linguistics.

\bibitem[{Li et~al.(2021)Li, Lin, Xiao, and Zhu}]{Li_Lin_Xiao_Zhu_2021}
Yanyang Li, Ye~Lin, Tong Xiao, and Jingbo Zhu. 2021.
\newblock \href {https://ojs.aaai.org/index.php/AAAI/article/view/17572} {An
  efficient transformer decoder with compressed sub-layers}.
\newblock \emph{Proceedings of the AAAI Conference on Artificial Intelligence},
  35(15):13315--13323.

\bibitem[{Liu et~al.(2020)Liu, Jiang, He, Chen, Liu, Gao, and Han}]{Liu2020On}
Liyuan Liu, Haoming Jiang, Pengcheng He, Weizhu Chen, Xiaodong Liu, Jianfeng
  Gao, and Jiawei Han. 2020.
\newblock \href {https://openreview.net/forum?id=rkgz2aEKDr} {On the variance
  of the adaptive learning rate and beyond}.
\newblock In \emph{International Conference on Learning Representations}.

\bibitem[{Post(2018)}]{post-2018-call}
Matt Post. 2018.
\newblock \href {https://doi.org/10.18653/v1/W18-6319} {A call for clarity in
  reporting {BLEU} scores}.
\newblock In \emph{Proceedings of the Third Conference on Machine Translation:
  Research Papers}, pages 186--191, Brussels, Belgium. Association for
  Computational Linguistics.

\bibitem[{Rei et~al.(2020)Rei, Stewart, Farinha, and
  Lavie}]{rei-etal-2020-comet}
Ricardo Rei, Craig Stewart, Ana~C Farinha, and Alon Lavie. 2020.
\newblock \href {https://doi.org/10.18653/v1/2020.emnlp-main.213} {{COMET}: A
  neural framework for {MT} evaluation}.
\newblock In \emph{Proceedings of the 2020 Conference on Empirical Methods in
  Natural Language Processing (EMNLP)}, pages 2685--2702, Online. Association
  for Computational Linguistics.

\bibitem[{Utiyama and Isahara(2007)}]{utiyama-isahara-2007-comparison}
Masao Utiyama and Hitoshi Isahara. 2007.
\newblock \href {https://aclanthology.org/N07-1061} {A comparison of pivot
  methods for phrase-based statistical machine translation}.
\newblock In \emph{Human Language Technologies 2007: The Conference of the
  North {A}merican Chapter of the Association for Computational Linguistics;
  Proceedings of the Main Conference}, pages 484--491, Rochester, New York.
  Association for Computational Linguistics.

\bibitem[{Vaswani et~al.(2017)Vaswani, Shazeer, Parmar, Uszkoreit, Jones,
  Gomez, Kaiser, and Polosukhin}]{NIPS2017_3f5ee243}
Ashish Vaswani, Noam Shazeer, Niki Parmar, Jakob Uszkoreit, Llion Jones,
  Aidan~N Gomez, \L~ukasz Kaiser, and Illia Polosukhin. 2017.
\newblock \href
  {https://proceedings.neurips.cc/paper/2017/file/3f5ee243547dee91fbd053c1c4a845aa-Paper.pdf}
  {Attention is all you need}.
\newblock In \emph{Advances in Neural Information Processing Systems},
  volume~30. Curran Associates, Inc.

\bibitem[{Wang et~al.(2020)Wang, Lipton, and
  Tsvetkov}]{wang-etal-2020-negative}
Zirui Wang, Zachary~C. Lipton, and Yulia Tsvetkov. 2020.
\newblock \href {https://doi.org/10.18653/v1/2020.emnlp-main.359} {On negative
  interference in multilingual models: Findings and a meta-learning treatment}.
\newblock In \emph{Proceedings of the 2020 Conference on Empirical Methods in
  Natural Language Processing (EMNLP)}, pages 4438--4450, Online. Association
  for Computational Linguistics.

\bibitem[{Wenzek et~al.(2021)Wenzek, Chaudhary, Fan, Gomez, Goyal, Jain, Kiela,
  Thrush, and Guzm{\'a}n}]{wenzek2021findings}
Guillaume Wenzek, Vishrav Chaudhary, Angela Fan, Sahir Gomez, Naman Goyal,
  Somya Jain, Douwe Kiela, Tristan Thrush, and Francisco Guzm{\'a}n. 2021.
\newblock \href {https://statmt.org/wmt21/pdf/2021.wmt-1.2.pdf} {Findings of
  the wmt 2021 shared task on large-scale multilingual machine translation}.
\newblock In \emph{Proceedings of the Sixth Conference on Machine Translation},
  pages 89--99.

\bibitem[{Yang et~al.(2021{\natexlab{a}})Yang, Ma, Huang, Zhang, Dong, Huang,
  Muzio, Singhal, Hassan, Song, and Wei}]{yang-etal-2021-multilingual-machine}
Jian Yang, Shuming Ma, Haoyang Huang, Dongdong Zhang, Li~Dong, Shaohan Huang,
  Alexandre Muzio, Saksham Singhal, Hany Hassan, Xia Song, and Furu Wei.
  2021{\natexlab{a}}.
\newblock \href {https://aclanthology.org/2021.wmt-1.54} {Multilingual machine
  translation systems from {M}icrosoft for {WMT}21 shared task}.
\newblock In \emph{Proceedings of the Sixth Conference on Machine Translation},
  pages 446--455, Online. Association for Computational Linguistics.

\bibitem[{Yang et~al.(2021{\natexlab{b}})Yang, Eriguchi, Muzio, Tadepalli, Lee,
  and Hassan}]{yang-etal-2021-improving-multilingual}
Yilin Yang, Akiko Eriguchi, Alexandre Muzio, Prasad Tadepalli, Stefan Lee, and
  Hany Hassan. 2021{\natexlab{b}}.
\newblock \href {https://doi.org/10.18653/v1/2021.emnlp-main.578} {Improving
  multilingual translation by representation and gradient regularization}.
\newblock In \emph{Proceedings of the 2021 Conference on Empirical Methods in
  Natural Language Processing}, pages 7266--7279, Online and Punta Cana,
  Dominican Republic. Association for Computational Linguistics.

\bibitem[{Zhang et~al.(2020)Zhang, Williams, Titov, and
  Sennrich}]{zhang-etal-2020-improving}
Biao Zhang, Philip Williams, Ivan Titov, and Rico Sennrich. 2020.
\newblock \href {https://doi.org/10.18653/v1/2020.acl-main.148} {Improving
  massively multilingual neural machine translation and zero-shot translation}.
\newblock In \emph{Proceedings of the 58th Annual Meeting of the Association
  for Computational Linguistics}, pages 1628--1639, Online. Association for
  Computational Linguistics.

\bibitem[{Zoph and Knight(2016)}]{zoph-knight-2016-multi}
Barret Zoph and Kevin Knight. 2016.
\newblock \href {https://doi.org/10.18653/v1/N16-1004} {Multi-source neural
  translation}.
\newblock In \emph{Proceedings of the 2016 Conference of the North {A}merican
  Chapter of the Association for Computational Linguistics: Human Language
  Technologies}, pages 30--34, San Diego, California. Association for
  Computational Linguistics.

\end{thebibliography}
\end{document}